\documentclass[preprint,12pt]{elsarticle}

\usepackage{amssymb}
\usepackage{amsmath}
\usepackage{color}
\usepackage{booktabs}
\usepackage{subcaption}
\usepackage{changes}
\biboptions{sort&compress}

\journal{Knowledge-Based Systems}

\begin{document}

\begin{frontmatter}

\title{
DnSwin: Toward Real-World Denoising via a Continuous Wavelet Sliding Transformer
}

\author[inst1]{Hao Li}
\author[inst1]{Zhijing Yang}

\author[inst2]{Xiaobin Hong \corref{cor1}}
\author[inst3]{Ziying Zhao}
\author[inst1]{Junyang Chen}
\author[inst1]{Yukai Shi}
\author[inst4]{Jinshan Pan}

\affiliation[inst1]{organization={School of Information Engineering, Guangdong University of Technology},
            city={Guangzhou},
            postcode={510006}, 
            country={China}}

\affiliation[inst2]{organization={School of Mechanical Automotive Engineering, South China University of Technology},
            city={Guangzhou},
            postcode={510641}, 
            country={China}}
            
\affiliation[inst3]{organization={School of Computer Science and Engineering, Sun Yat-sen University},
            city={Guangzhou},
            country={China}}

\affiliation[inst4]{organization={School of Computer Science and Engineering, Nanjing University of Science and Technology},
            city={Nanjing},
            country={China}}

\cortext[cor1]{Corresponding author}

\begin{abstract}
Real-world image denoising is a practical image restoration problem that aims to obtain clean images from in-the-wild noisy inputs. Recently, the Vision Transformer (ViT) has exhibited a strong ability to capture long-range dependencies, and many researchers have attempted to apply the ViT to image denoising tasks. However, a real-world image is an isolated frame that makes the ViT build long-range dependencies based on the internal patches, which divides images into patches, disarranges noise patterns and damages gradient continuity. In this article, we propose to resolve this issue by using a continuous Wavelet Sliding-Transformer that builds frequency correspondences under real-world scenes, called DnSwin. Specifically, we first extract the bottom features from noisy input images by using a convolutional neural network (CNN) encoder. The key to DnSwin is to extract high-frequency and low-frequency information from the observed features and build frequency dependencies. To this end, we propose a Wavelet Sliding-Window Transformer (WSWT) that utilizes the discrete wavelet transform (DWT), self-attention and the inverse DWT (IDWT) to extract deep features. Finally, we reconstruct the deep features into denoised images using a CNN decoder. Both quantitative and qualitative evaluations conducted on real-world denoising benchmarks demonstrate that the proposed DnSwin performs favorably against the state-of-the-art methods.
\end{abstract}


\end{frontmatter}

\section{Introduction}

Image denoising is a fundamental topic in low-level vision fields and is widely applied in image restoration~\cite{KBS1}, super-resolution~\cite{KBS2,li2022real}, optical flow~\cite{flow}, \textit{etc.}. It also plays an important role in the preprocessing steps of high-level computer vision tasks, such as object detection and image segmentation~\cite{connecting,sod}. The purpose of image denoising is to restore a latent clean image $x$ from its noisy observation $y$. Traditional image denoising can be formulated as follows:
\begin{equation}
y = x + n.
\end{equation}
where $n$ is the additive noise that is generally modeled as zero-mean white Gaussian noise with a standard deviation of $\sigma$. To address image degradation problems, many traditional image recovery models and learning-based methods have been developed in recent decades.

\begin{figure}[ht]
    \centering
    \includegraphics[width=0.9\linewidth]{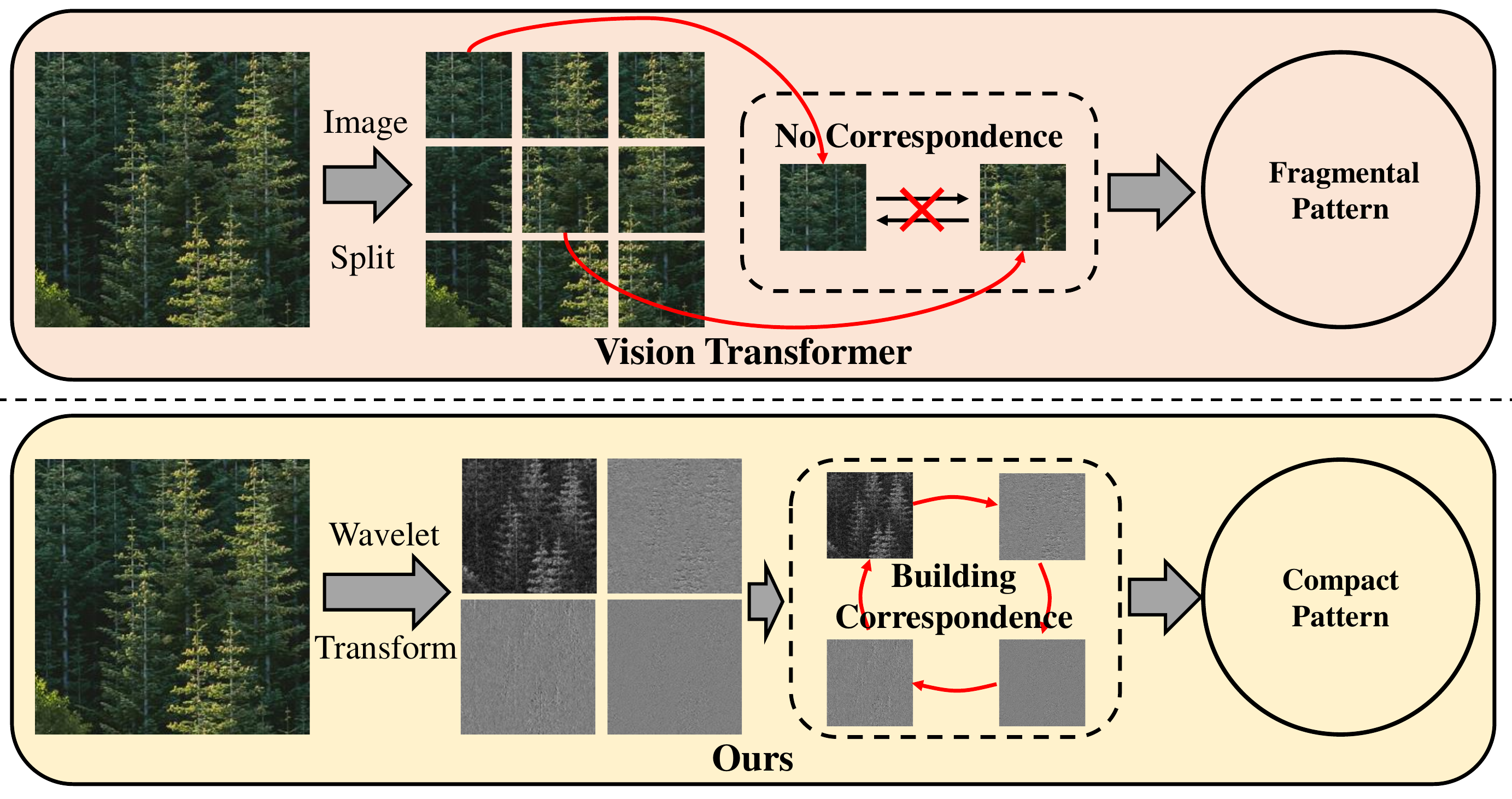}
    \caption{The tokenization strategy 
in the ViT roughly divides the input image into patches and breaks the image continuity. We address the \textbf{long-range correspondence} on a compact yet intact representation via a wavelet sliding-transformer.}
    \label{fig:VandW}
\end{figure}

\begin{figure}[ht]
    \centering
    \includegraphics[width=0.8\linewidth]{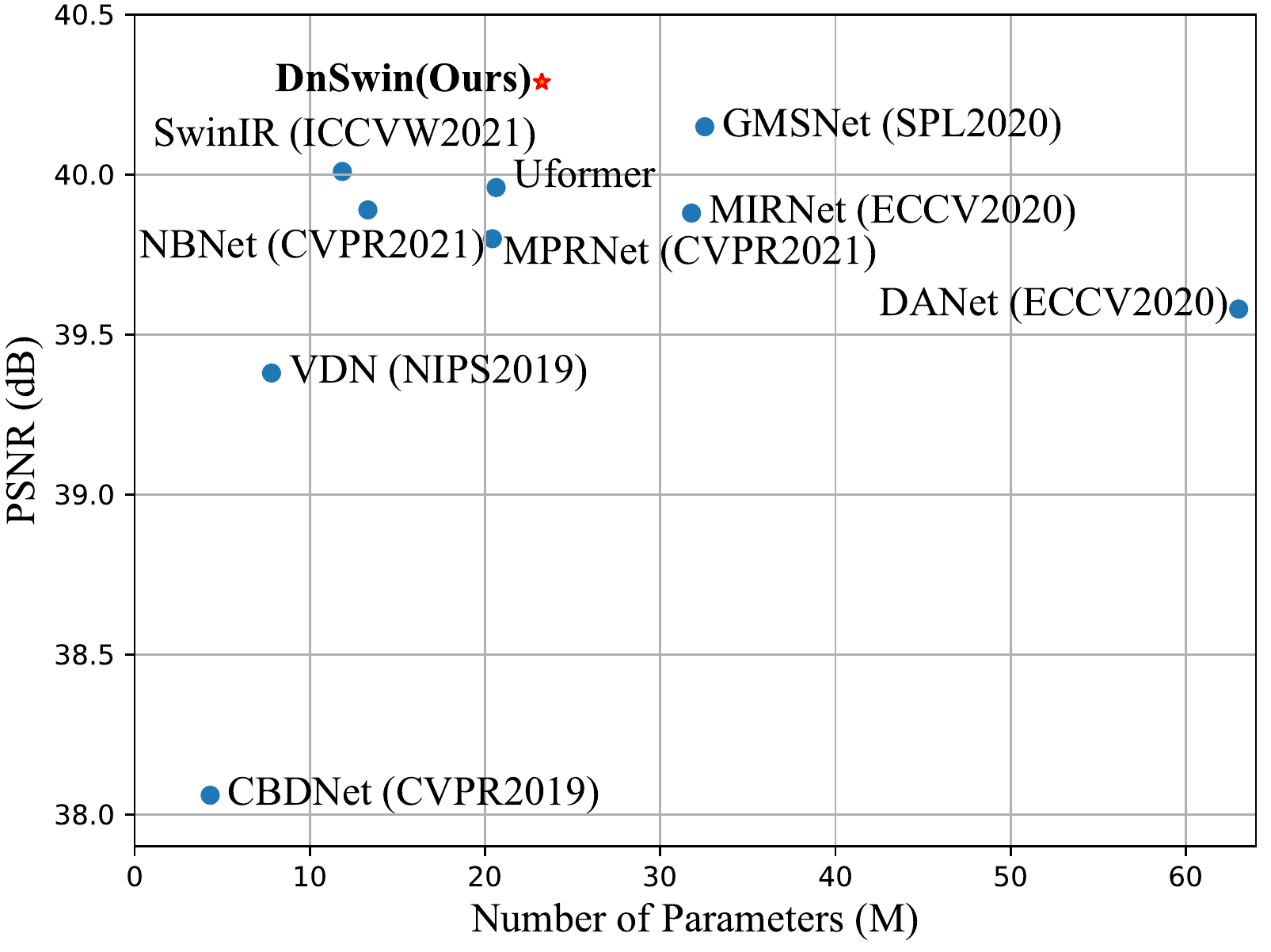}
    \caption{PSNR results vs. the total number of parameters for different methods in real-world image denoising tasks conducted on the DND~\cite{dnd} benchmark.}
    \label{fig:Para}
\end{figure}

\begin{figure}[ht]
    \centering
    \includegraphics[width=\linewidth]{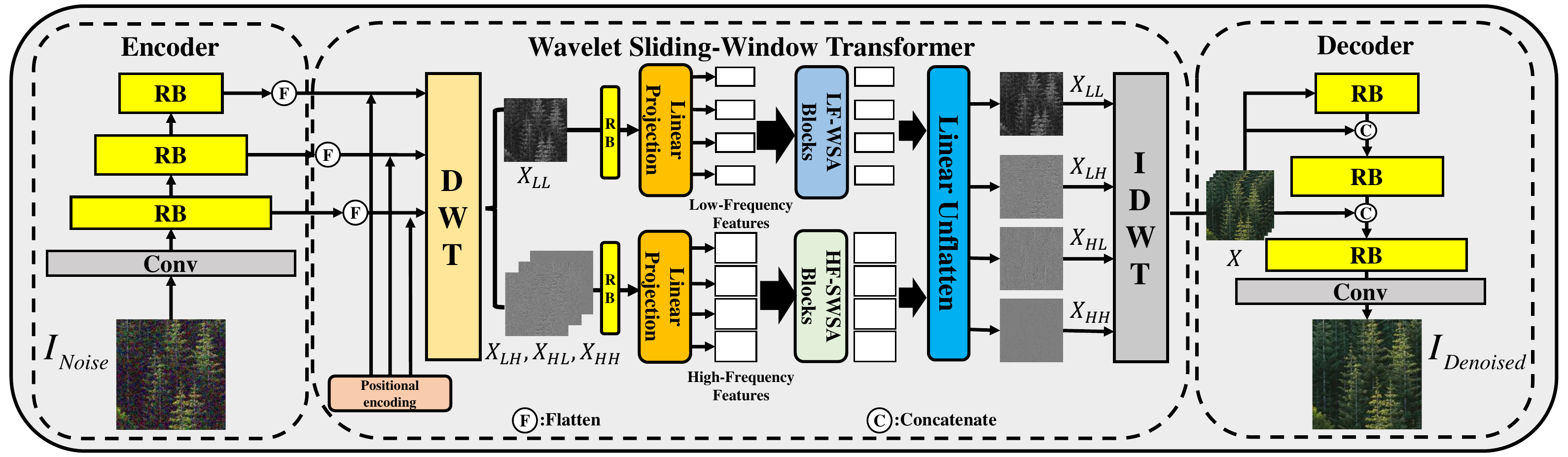}
    \caption{The overall architecture of DnSwin consists of an encoder module, a Wavelet Sliding-Window Transformer (WSWT) and a decoder module.}
    \label{fig:network}
\end{figure}

Traditional model-based solutions try to explore the prior image knowledge to regularize this essentially ill-posed degradation processing problem from a Bayesian viewpoint. In recent decades, various methods have been proposed by learning prior knowledge from nonlocal self-similarity~\cite{NLB}, sparse representation~\cite{sparse_representation}, and total variation~\cite{total_variation} models. Although the above methods have produced satisfactory results, they usually suffer from utilizing too many manually selected features to constrain images, which makes their optimization process time-consuming and limits their practicality.

Compared with model-based denoising methods, the existing learning-based restoration methods~\cite{CBDNet,vdn,tian2021designing,li2021joint} address the above problem more efficiently and obtain charming visual quality when handing images with synthetic noise, such as additive white Gaussian noise (AWGN). However, real-world image denoising remains a challenging task, as the processing steps within camera systems, such as demosaicing, gamma correction and compression, are diversely constructed. In recent years, researchers have tried to build convolutional neural network (CNN)-based denoisers to handle real-world noisy photographs by modeling real noise~\cite{CBDNet,unprocessing}, obtaining noisy and noise-free image pairs~\cite{SIDD_2018_CVPR,dnd}, and applying unsupervised/self-supervised learning strategies~\cite{DBSN,neighbor2neighbor}. Typically, CNN-based real-world denoising approaches~\cite{danet,nbnet} choose `UNet'~\cite{unet} as their backbone architecture. Nevertheless, the reduplicative downsampling operations in UNet cause serious damage to high-frequency information, which is consequential for image content preservation.

To address this problem, several recent works~\cite{ttsr,pipt} have attempted to utilize the vision transformer (ViT) for image restoration tasks. However, two challenges are involved in using the ViT for image restoration. First, the standard ViT structure~\cite{vit} applies a global attention mechanism between all tokens. As shown in \figurename~\ref{fig:VandW}, the ViT exhibits a strong ability to capture long-range dependencies. However, in a real-world denoising task, the real-world image is an isolated frame for which the ViT must build dependencies on the fragmental patches. A ViT-based denoiser realizes tokenization by cutting a large-size image into patches, which breaks the noise pattern. Second, previous works~\cite{cvt,localvit} showed that the ViT has a limitation in capturing local dependencies, which is essential for image restoration tasks.

To address these issues, we propose a novel hybrid framework for image denoising, which is depicted in \figurename~\ref{fig:network}. DnSwin consists of three encoder stages, three decoder stages and a wavelet sliding-window transformer (WSWT). In our model, we first utilize a CNN encoder to encode bottom features derived from noisy input images. As the discrete wavelet transform (DWT) can preserve the information contained in different frequency bands, which is important when recovering images, we apply the DWT to achieve the separation of low-frequency features and high-frequency features. Furthermore, we introduce low-frequency window-based self-attention (LF-WSA) and high-frequency sliding window-based self-attention (HF-SWSA) to build frequency correspondences on compact yet intact representations. Subsequently, we integrate the multifrequency subbands into features using the inverse DWT (IDWT). Finally, we reconstruct the features into denoised images using a CNN decoder.

Extensive experiments are conducted to evaluate DnSwin on real-world denoising tasks. The results obtained on four real-world denoising benchmarks show that DnSwin achieves state-of-the-art denoising performance on real-world scenes. Especially on the Darmstadt Noise Dataset (DND)~\cite{dnd} benchmark, as shown in \figurename~\ref{fig:Para}, our proposed DnSwin achieves obvious improvements over other state-of-the-art methods while utilizing suitable parameters. Generally, our contributions can be summarized as follows.
\begin{itemize}
\item We propose a wavelet ViT architecture for real-world image denoising. The proposed model captures long-range correspondences along the frequency subbands on compact yet intact image content. This strategy disentangles structural information from high- and low-frequency correspondences and provides a better interpretation for the ViT-based denoiser.
\item We propose an LF-WSA block and an HF-SWSA block to make DnSwin intently exact features with different frequencies, especially the noise features in high frequency.
\item Experimental results demonstrate that our model achieves state-of-the-art performance on real-world benchmarks in terms of peak signal-to-noise ratio (PSNR) and structural similarity index measure (SSIM).
\end{itemize}

The remaining parts of this paper are organized as follows. Section~\ref{section: Related} briefly surveys the advances achieved in deep denoising models dealing with AWGN and real-world noise and the progress made regarding the ViT. Section~\ref{section: methodology} presents the network architecture and detailed explanations of the LF-WSA and HF-SWSA. In Section~\ref{section: Experiment}, we perform experiments to verify the effectiveness and efficiency of DnSwin and compare it with existing state-of-the-art methods. Finally, we conclude our work and propose some future research directions in Section~\ref{section: Conclusion}.

\section{Related Work}
\label{section: Related}
\textbf{Gaussian Denoising.} As a fundamental low-level image processing challenge, image denoising has been intensively studied for decades. Most traditional denoising methods can be interpreted within the maximum a posteriori (MAP) framework, e.g., total variation~\cite{total_variation}, sparse representation~\cite{sparse_representation}, wavelet coring~\cite{wavelet_coring} and nonlocal self-similarity~\cite{NLB}. These methods learn prior image model parameters via recognition and compact unrolled inference and introduce model-guided discriminatory learning. However, these methods tend to oversmooth image details. To this end, Mahdaoui et al.~\cite{s22062199} proposed a compressed sensing (CS) reconstruction method that combines total variation regularization and the nonlocal self-similarity constraint to achieve significant improvements in terms of denoising efficiency and visual quality.

Benefiting from the success of deep learning, many researchers have focused on the exploration of CNN structures. Advanced architecture designs have achieved very competitive performance in comparison with traditional methods. For instance, Burger et al.~\cite{bm3d} trained a plain multi-layer perceptron (MLP) with a large synthetic noise dataset, achieving comparable results to those of block matching and 3D filtering (BM3D). Chen et al.~\cite{tnrd} proposed a trainable nonlinear reaction diffusion (TNRD) approach to perform truncated gradient descent inference. Zhang et al.~\cite{zhang2017beyond} further proposed the denoising CNN (DnCNN), which forms a model by utilizing batch normalization (BN)~\cite{ioffe2015batch} and residual connection~\cite{he2016deep}. Mao et al.~\cite{mao2016image} proposed the residual encoder-decoder network (REDNet), a deep fully convolutional encoding-decoding network with symmetric skip connections.

Driven by the advances in self-attention~\cite{attention}, many approaches try to extract the global features of images by using a self-attention mechanism to improve the resulting visual effects. Mei et al.~\cite{mei2020pyramid} proposed a novel pyramid attention module for image restoration, which can capture long-range feature correspondences from a multi-scale feature pyramid. Zhang et al.~\cite{zhang2019residual} designed nonlocal attention blocks to capture global information and pay more attention to the challenging parts of images. However, this method leads to high memory occupation and time consumption levels.

Nevertheless, the above learning-based denoising approaches are all trained with synthetic noisy datasets, so they can only perform well when processing images with AWGN but significantly degrade when applied to real-world images~\cite{shi2022sr}. Inspired by the aforementioned works, we propose a novel hybrid network architecture that inherits the superiority of a CNN and the ViT and demonstrates a stronger generalization ability.

\textbf{Real-World Noisy Images.} For real-world noisy images, the noise model is indeed distinct from AWGN. Actually, real noise is sophisticated, and the camera image signal processing (ISP) pipeline further increases its complexity. As a remedy, existing deep denoisers for handling real-world noisy images are usually trained either by exploiting realistic noise models to synthesize noisy images or acquiring real paired noisy and noise-free images. The convolutional blind denoising network (CBDNet)~\cite{CBDNet} utilizes a realistic noise model including heterogeneous Gaussian and ISP pipelines; it is composed of two subnetworks, including a noise estimation module and a nonblind denoising module. Chen et al. proposed GCBD~\cite{GCBD} to extract a smaller image with a clear background from a noisy image and then incorporated a generative adversarial network (GAN) to generate more fake noise samples for training the denoising CNN. RIDNet~\cite{RIDNet} uses a single-state denoising network with feature attention for real-world noise reduction. GRDN~\cite{GRDN} was the champion of the sRGB track at NTIRE 2019~\cite{ntire2019}; it incorporates groups of residual dense blocks (GRDBs) and convolutional block attention modules (CBAMs) for real-world image denoising.

However, the above approaches ignore the correspondence between multi-frequency features and fail to build local and global contexts; therefore, the complex noise information inside real-world images cannot be effectively removed. To this end, DnSwin utilizes a WSWT to decompose, extract, and integrate multi-frequency features, achieving state-of-the-art results on real-world denoising benchmarks.

\textbf{Vision Transformer.} Transformer~\cite{attention} has achieved significant performance in natural language processing (NLP). Compared with a CNN, the multi-head self-attention (MSA) and feedforward MLP layer in a transformer are naturally good at capturing long-range dependencies between words via global self-attention. Due to the success of Transformer in NLP, many attempts have been made to explore the potential of transformers in computer vision tasks. The pioneering work on the ViT~\cite{image_transformer} obtained excellent results compared to those of state-of-the-art CNNs on image classification tasks by dividing images into $16 \times 16$ image patches and then stretching them into one-dimensional vectors that were fed into Transformer blocks. Chen et al.~\cite{transunet} proposed TransUnet, a hybrid framework using the ViT and UNet based on convolutional operations for medical image segmentation. Because of the limitation of the ViT in capturing local contexts, several works have tried to add convolutional layers to the MSA module~\cite{cvt,convit} and the feedforward network~\cite{incorporating,localvit}. Liu et al.~\cite{Swin} proposed the Swin Transformer, which significantly reduces the computational cost of the ViT by using shifted window-based multi-head attention mechanisms.

In addition to image classification and segmentation, Transformer has also exhibited potential in image restoration tasks. Yang et al.~\cite{ttsr} first proposed the texture Transformer network for image super-resolution (TTSR) and incorporated the ViT-based architecture into image restoration tasks. They used a cross-scale feature integration module to stack a texture transformer, thereby achieving a more powerful feature transformer. IPT~\cite{pipt} exploits the potential of pretraining and transfer learning with a shared ViT-based backbone. To address multiple image restoration tasks (e.g., super-resolution, denoising, and deraining), it employs each head and each tail separately for each image restoration task. However, the above approaches still need large-scale reference data for training. In contrast, DnSwin is able to achieve surprising performance with limited training samples by incorporating a wavelet CNN and the ViT into a combination-based architecture.

\section{Methodology}
\label{section: methodology}

\begin{figure}[ht]
    \centering
    \includegraphics[width=0.8\linewidth]{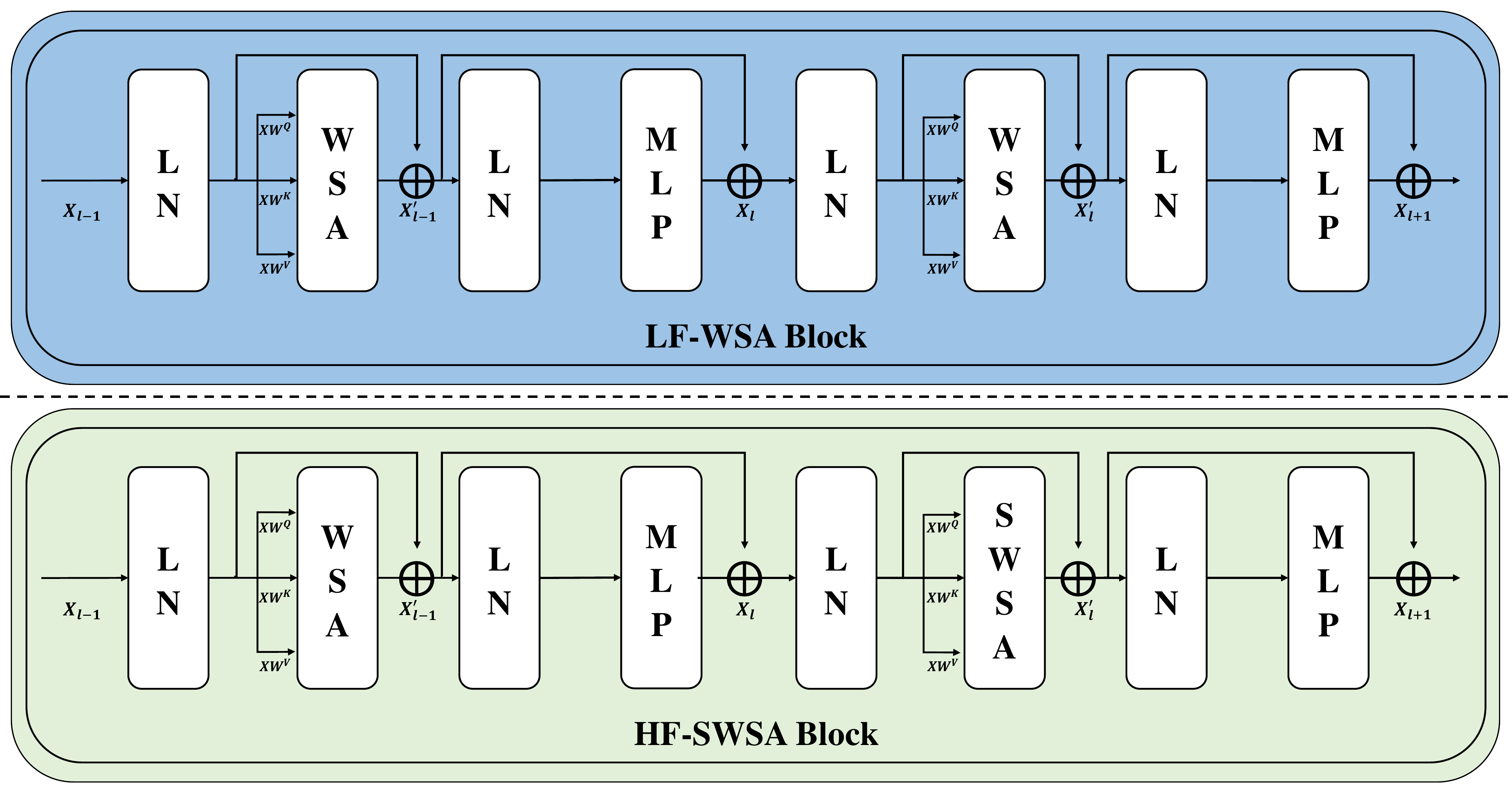}
    \caption{The details of the LF-WSA block and the HF-SWSA block.}
    \label{fig:LF-WSA}
\end{figure}

In this section, we present our real-world image denoising network based on DnSwin. We first briefly describe the overall architecture details of DnSwin. Then, we provide more detailed explanations about bottom feature acquisition, the WSWT with three submodules (i.e., wavelet feature decomposition, frequency correspondence building and multi-frequency integration) and reconstruction.

\subsection{Overall Architecture Details}
As shown in \figurename~\ref{fig:network}, DnSwin consists of three components: an encoder module, a WSWT and a decoder module. Specifically, the feature encoder module uses convolutional layers to extract the bottom features:
\begin{equation}
    F^{i}_{b}=H_ {E}(I_{noisy}),
\end{equation}
where $F^{i}_{b}$ denotes an extracted bottom feature, $i$ is the index of the layer, $H_{E}$ is the function of the feature encoder, and $I_{noisy}$ refers to the noisy image. Then, we use the WSWT to separate information with different frequencies from the bottom feature and build the frequency dependencies:
\begin{equation}
    F^{i}_{d}=H_{W}(F^{i}_{b}),
\end{equation}
where $H_{W}$ is the function of the WSWT and $F^{i}_{d}$ denotes the deep feature extracted by the WSWT. The feature decoder uses convolution layers and a concatenation operator to fuse hierarchical features and reconstruct the image:
\begin{equation}
    F^{i-1}_{r}=H_{D}(Concat(F^{i}_{d}\uparrow,F^{i-1}_{d})),
\end{equation}
where $F^{i}_{r}$ denotes the reconstruction feature obtained from the decoder (when $i=0$, $F^{0}_{r}=I_{Denoised}$ refers to the denoised image), $H_{D}$ is the function of the feature decoder, and $\uparrow$ is the upsampling operation.

\subsection{Bottom Feature Acquisition}
We utilize the encoder to execute bottom feature encoding for noisy input images. The feature encoder contains a convolutional layer and three stages of residual blocks (RBs). Each RB consists of three convolutional layers and two leaky ReLU layers. Specifically, given a noisy image $I_{noisy} \in \mathbb{R}^{3 \times H \times W}$ ($H$ and $W$ are the input image height and width, respectively), we first apply a $3\times3$ convolutional layer to extract features $\mathbf{X}^{0} \in \mathbb{R}^{C \times H \times W}$, where $C$ denotes the number of output channels. Next, the feature maps $\mathbf{X}^{l}$ produced by the RBs~\cite{he2016deep} at different stages can be formally expressed as:

\begin{equation}
\begin{aligned}
    &\mathbf{X}^{0}  = \operatorname{Conv}(I_{noisy}),\\
    &\mathbf{X}^{i}  = H_{RB} (\mathbf{X}^{i-1})\downarrow .\\
\end{aligned}
\label{equ:encoder}
\end{equation}
where $\operatorname{Conv}$ is the convolutional operation, $\{\mathbf{X}^{1},\mathbf{X}^{2},\dots,\mathbf{X}^{i}\} \in \mathbb{R}^{2^{i}C \times \frac{H}{2^{i}} \times \frac{W}{2^{i}}}$ denotes the output feature maps of the $i$-th stage, and $H_{RB}(\cdot)$ is the function of the RBs. In the downsampling layer $\downarrow$, we use a $4 \times 4$ convolution with a stride of 2, and the number of feature maps is doubled to reduce the information loss incurred during each downsampling step.

\subsection{Wavelet Sliding-Window Transformer}
Since the typical convolutional operator has a local receptive field, the feature encoder fails to capture the long-range dependencies of pixels. To this end, we propose the Wavelet Sliding-Window Transformer, which introduces a multi-frequency self-attention mechanism for long-range contextual extraction and correspondence modeling.
\subsubsection{Wavelet Feature Decomposition}
The DWT performs signal decomposition with attractive time-frequency localization properties that have the potential to decompose feature maps into different frequencies for further processing. More specifically, the DWT decomposes an image into a low-frequency subband and three high-frequency subbands. The low-frequency subband contains detailed information, while the high-frequency subbands contain structural information. Given the feature maps $\mathbf{X}=\{\mathbf{X}^{1},\mathbf{X}^{2},\dots,\mathbf{X}^{i}\}$ obtained by feature acquisition, we first flatten them into patches, apply learnable positional encoding for spatial information modeling, and then use the DWT to decompose the patches into different frequency intervals by the following equations:
\begin{equation}
\left\{\begin{matrix}
 \mathbf{X}_{LL} = (\mathbf{f}_{LL} * \mathbf{X}) \downarrow_{2}, \\ 
 \mathbf{X}_{LH} = (\mathbf{f}_{LH} * \mathbf{X}) \downarrow_{2}, \\
 \mathbf{X}_{HL} = (\mathbf{f}_{HL} * \mathbf{X}) \downarrow_{2}, \\
 \mathbf{X}_{HH} = (\mathbf{f}_{HH} * \mathbf{X}) \downarrow_{2}, \\
\end{matrix}\right.
\label{equ:DWT}
\end{equation}
where the subbands $\{\mathbf{X}_{LL}, \mathbf{X}_{LH}, \mathbf{X}_{HL}, \mathbf{X}_{HH}\}$ denote the approximation, horizontal detail, vertical detail and diagonal detail of the corresponding feature obtained using the Harr wavelet transform, respectively. $*$ is the convolution operator, and $\downarrow_{2}$ is the standard downsampling operator with a factor of 2. The low-pass filter $\mathbf{f}_{LL}$ and three typical high-pass filters $\mathbf{f}_{LH}$, $\mathbf{f}_{HL}$, and $\mathbf{f}_{HH}$ are defined as:
\begin{equation}
\begin{array}{l}
\mathbf{f}_{L L}=\left[\begin{array}{ll}
1 & 1 \\
1 & 1
\end{array}\right], \quad \mathbf{f}_{L H}=\left[\begin{array}{cc}
-1 & -1 \\
1 & 1
\end{array}\right], \\
\mathbf{f}_{H L}=\left[\begin{array}{ll}
-1 & 1 \\
-1 & 1
\end{array}\right], \quad \mathbf{f}_{H H}=\left[\begin{array}{cc}
1 & -1 \\
-1 & 1
\end{array}\right] .
\end{array}
\end{equation}

After we obtain $\{\mathbf{X}_{LL},\mathbf{X}_{LH},\mathbf{X}_{HL},\mathbf{X}_{HH}\}$, we divide them into a low-frequency subband $\mathbf{X}_{LL}$ and three high-frequency subbands $\{\mathbf{X}_{LH},\mathbf{X}_{HL},\mathbf{X}_{HH}\}$ and use RBs to extract the local features of the different frequency subbands.

To build correspondences between the different frequency subbands, we use LF-WSA and HF-SWSA to extract global features and capture long-range dependencies.

\subsubsection{Building Frequency Correspondences via Sliding-Window}
The details of LF-WSA and HF-SWSA are shown in \figurename~\ref{fig:LF-WSA}; they benefit from the self-attention mechanism in the transformer to capture long-range dependencies and build frequency correspondences between all subbands.

As shown in \figurename~\ref{fig:LF-WSA}, each HF-SWSA block consists of a layer normalization (LN)~\cite{layernorm} layer, a multi-head windows-based self-attention (WSA) unit, a mult-ihead sliding window-based self-attention (SWSA) unit and the 2-layer MLP with the GELU non-linearity activation function; the LF-WSA block is modified from the HF-SWSA block by replacing an SWSA unit with a WSA unit. The computation of the HF-SWSA block is represented as:
\begin{equation}
\begin{array}{l}
\mathbf{X}^{'}_{l}=\operatorname{WSA}\left(\mathrm{LN}\left(\mathbf{X}_{l-1}\right)\right)+\mathbf{X}_{l-1}, \\
\mathbf{X}_{l}=\operatorname{MLP}\left(\mathrm{LN}\left(\mathbf{X}^{'}_{l}\right)\right)+\mathbf{X}^{'}_{l}, \\
\mathbf{X}^{'}_{l+1}=\operatorname{SWSA}\left(\mathrm{LN}\left(\mathbf{X}_{l}\right)\right)+\mathbf{X}_{l}, \\
\mathbf{X}_{l+1}=\operatorname{MLP}\left(\mathrm{LN}\left(\mathbf{X}^{'}_{l+1}\right)\right)+\mathbf{X}^{'}_{l+1}.
\end{array}
\label{equ:WSW}
\end{equation}
where $\mathbf{X}^{'}_{l}$ and $\mathbf{X}_{l}$ represent the outputs of the WSA and SWSA units and the MLP module in the $l$-th layer, respectively. Next, we introduce the details of the WSA and SWSA.

\emph{\textbf{Windows based Self Attention.}} Given a feature map $\mathbf{X} \in \mathbb{R}^{C \times H \times W}$, we first split the feature map into nonoverlapping windows with a window size of $M \times M$ and obtain the transposed features $\{\mathbf{X}^{1},\mathbf{X}^{2},\dots,\mathbf{X}^{N}\} \in \mathbb{R}^{M^{2} \times C},~ N = H W/M^{2}$ from each window. Then, we apply a self-attention mechanism on $\mathbf{X}^{N}$. Suppose that the number of heads is $h$ and that the head dimensionality is $d = C/h$. The workflow of WSA can be formulated as:
\begin{equation}
\normalsize
\begin{array}{l}
     \mathbf{X} = \{\mathbf{X}^{1},\mathbf{X}^{2}, \dots, \mathbf{X}^{N} \}, ~ N = H W/M^{2},  \\
     \mathbf{Y}^{n}_{k} = \operatorname{Attention}(\mathbf{X}^{n}\mathbf{W}^{Q}_{k},\mathbf{X}^{n}\mathbf{W}^{K}_{k},\mathbf{X}^{n}\mathbf{W}^{V}_{k}), ~n = 1,\dots,N, \\
     \hat{\mathbf{X}_{k}}=\{\mathbf{Y}^{1}_{k},\mathbf{Y}^{2}_{k},\dots,\mathbf{Y}^{N}_{k} \},
\end{array}
\label{equ:Win-SA}
\end{equation}
where $\mathbf{W}^{Q}_{k},\mathbf{W}^{K}_{k},\mathbf{W}^{V}_{k} \in \mathbb{R}^{C \times d}$ denote the query, key and value window-based projection matrices for the $k$-th head, respectively. The outputs for all heads $\{\hat{\mathbf{X}_{1}},\hat{\mathbf{X}_{2}},\dots,\hat{\mathbf{X}_{k}}\}$ are concatenated and linearly projected to obtain the final result. Following previous works, the self-attention mechanism can be formulated as:
\begin{equation}
\begin{split}
    &\operatorname{Attention}(\mathbf{X}\mathbf{W}^{Q},\mathbf{X}\mathbf{W}^{K},\mathbf{X}\mathbf{W}^{V}) \\ =&\operatorname{SoftMax}(\frac{(\mathbf{X}\mathbf{W}^{Q})(\mathbf{X}\mathbf{W}^{K})^{T}}{\sqrt{d}}+B)(\mathbf{X}\mathbf{W}^{V}),
\end{split}
\label{equ:attention}
\end{equation}
where $B$ denotes the relative position bias, and the values in $B$ are taken from the bias matrix $\hat{B} \in \mathbb{R}^{(2M-1) \times (2M-1)}$. 

\begin{figure}[ht]
	\begin{center}
		\includegraphics[width=0.7\linewidth]{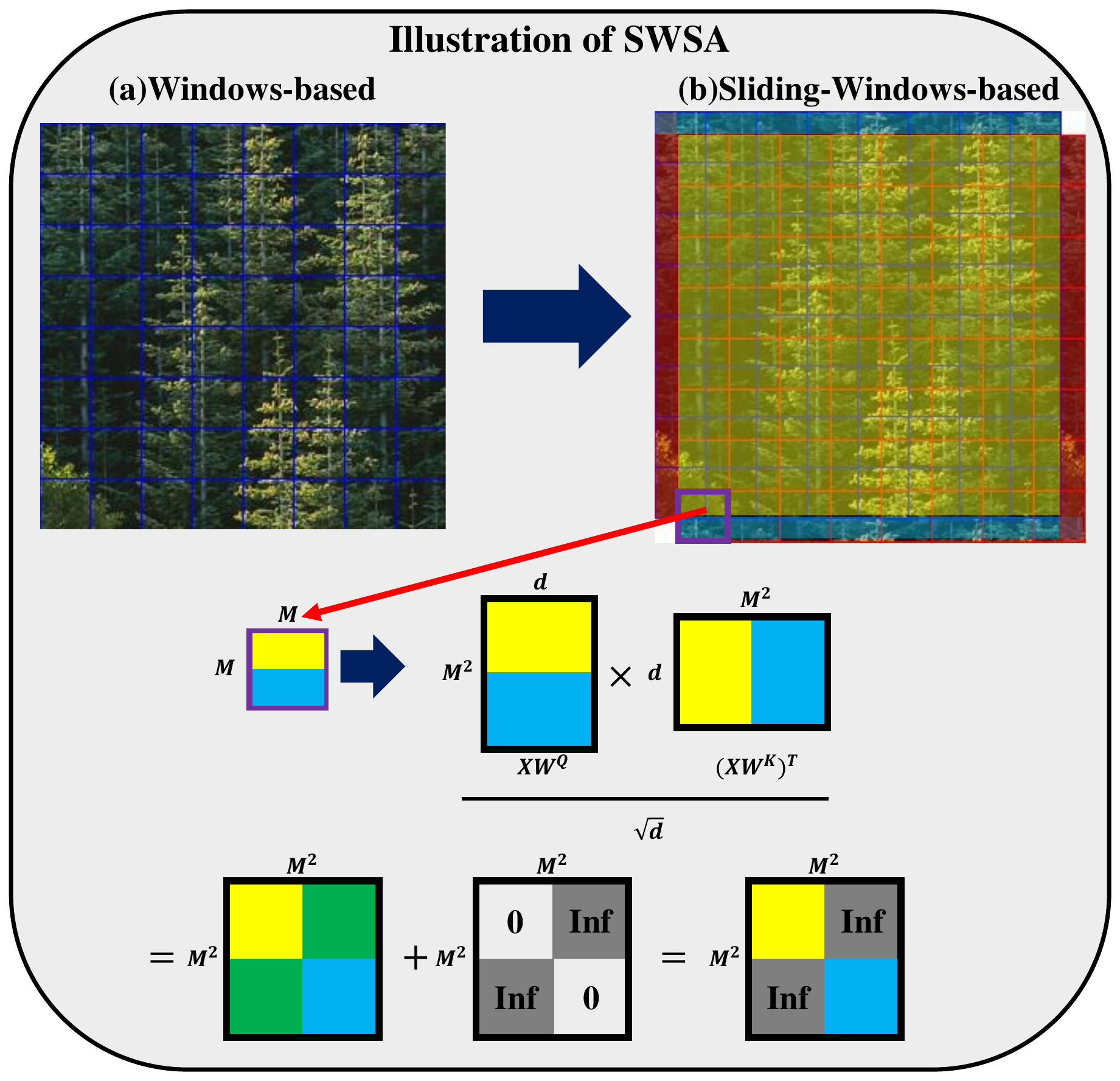}
	\end{center}
	\caption{Illustration of SWSA.}
	\label{fig:SWSA}
\end{figure}

As shown in \figurename~\ref{fig:network}, for the low-frequency subband, we use LF-WSA blocks to extract features to reduce the computational cost. For high-frequency subbands, we replace the LF-WSA blocks with HF-SWSA blocks to further extract the noise features, as noise belongs to the high-frequency features.

\emph{\textbf{Sliding-Windows based Self Attention.}} The WSA mechanism performs self-attention within each window and lacks cross-window connectivity, which limits its modeling capabilities. Based on WSA, we introduce a sliding mechanism to build connections across windows called the SWSA mechanism.

As shown in \figurename~\ref{fig:SWSA}(a), given a feature map $\mathbf{X} \in \mathbb{R}^{C \times H \times W}$, we use the same window partitioning strategy as that of WSA. Then, we slide all windows to the bottom right by $(\lfloor \frac{M}{2} \rfloor,\lfloor \frac{M}{2} \rfloor)$ from the top-left pixel, through which the red part on the left side and the blue part on top in the feature move to the right and to the bottom, respectively, as shown in \figurename~\ref{fig:SWSA}(b). The SWSA process is divided into the following parts. 1) The sub-windows consisting entirely of yellow overlapping areas are similar to those in Eq.~\eqref{equ:attention}. 2) The sub-windows consisting of areas with different colors are illustrated at the bottom of \figurename~\ref{fig:SWSA}. Specifically, the color area ({\textcolor[RGB]{238,201,0}{yellow}}, {\color{blue}blue}) in the feature remains the same color after applying the self-attention mechanism to the purple feature. In contrast, the {\color{green}green} area obtained from the matrix calculation of the {\textcolor[RGB]{238,201,0}{yellow}} and {\color{blue}blue} areas is meaningless, so we apply a mask mechanism to it to restrict the self-attention computation within each sub-window.

\emph{\textbf{Sliding Mechanism.}} Benefiting from the sliding mechanism, three high-frequency subbands build correspondences that can effectively extract different noise features at high frequencies. More specifically, given the low-frequency feature map $\mathbf{X}_{LL}$ and high-frequency feature maps $\{\mathbf{X}_{LH},\mathbf{X}_{HL},\mathbf{X}_{HH} \}$, we extract deep features from the LF-WSA and HF-SWSA mechanisms as follows:
\begin{equation}
\begin{array}{c}
  \hat{\mathbf{X}_{LL}} = H_{LF}(\mathbf{X}_{LL}), \\
  \{\hat{\mathbf{X}_{LH}},\hat{\mathbf{X}_{HL}},\hat{\mathbf{X}_{HH}}\} = H_{HF}(\{\mathbf{X}_{LH},\mathbf{X}_{HL},\mathbf{X}_{HH}\}).
\end{array}
\label{equ:LFHF}
\end{equation}
where $H_{LF}(\cdot)$ and $H_{HF}(\cdot)$ are the LF-WSA and HF-SWSA blocks, respectively.

Furthermore, WSA can significantly reduce the computational cost compared with that of global self-attention. Given the feature map $\mathbf{X} \in \mathbb{R}^{C \times H \times W}$ and a window size of $M$, the computational complexity levels of the global MSA and WSA mechanisms are:
\begin{equation}
\begin{array}{ll}
     & O(\operatorname{MSA}) = 4HWC^2 + 2(HW)^2C, \\
     & O(\operatorname{WSA}) = 4HWC^2 + 2M^2HWC.
\end{array}
\end{equation}

\textbf{Difference from the Swin Transformer.} In the Swin Transformer~\cite{Swin}, tokenization is a necessary step, but it is inappropriate for large-size image denoising. If we simply apply the Swin Transformer to a real-world denoising task, the full-size image will be cut into fragmental patches, which undoubtedly breaks the image continuity and noise patterns. In the proposed DnSwin, we combine the wavelet operator with Transformer, which preserves compact yet intact feature representations. Moreover, we investigate a sliding transformer to build long-range correspondences along the frequency subband on the unbroken image content. This strategy disentangles the structural information from the high- and low-frequency correspondences to better interpret real-world image denoising.

\subsubsection{Multi-frequency Integration}
As shown in \figurename~\ref{fig:network}, since we design DnSwin as a hierarchical architecture, the LF-WSA blocks at the low-frequency feature map $\mathbf{X}_{LL}$ work on larger receptive fields and are sufficient for learning long-range dependencies. The blocks at the high-frequency feature maps $\{\mathbf{X}_{LH},\mathbf{X}_{HL},\mathbf{X}_{HH} \}$ learn the different noise features and build connections between the three high-frequency subbands.

Nevertheless, we need to build correspondences between the low-frequency subband and the high-frequency subbands. Therefore, we propose applying the IDWT to the integration of multi-frequency subbands. Specifically, we first obtain the outputs $\hat{\mathbf{X}_{LL}}$ and $\{\hat{\mathbf{X}_{LH}},\hat{\mathbf{X}_{HL}},\hat{\mathbf{X}_{HH}}\}$ of the LF-WSA blocks and HF-SWSA blocks, as described by Eq.~\eqref{equ:LFHF}; then, we unflatten and expand them to feature maps $\{\hat{\mathbf{X}_{LL}},\hat{\mathbf{X}_{LH}},\hat{\mathbf{X}_{HL}},\hat{\mathbf{X}_{HH}}\}$. Finally, the multi-frequency feature maps $\{\hat{\mathbf{X}_{LL}},\hat{\mathbf{X}_{LH}},\hat{\mathbf{X}_{HL}},\hat{\mathbf{X}_{HH}}\}$ are integrated by the IDWT, which can accurately reconstruct the original features from the wavelet subbands. The operation process of the IDWT is defined as follows:
\begin{equation}
\small
\left\{\begin{aligned}
&\hat{\mathbf{X}}(2i-1,2j-1)=(\hat{\mathbf{X}_{LL}}(i,j)-\hat{\mathbf{X}_{LH}}(i,j)\\
&\;\;\;\;\;\;\;\;\;\;\;\;\;\;\;\;\;\;\;\;\;\;\;\;\;\;\;\;\;-\hat{\mathbf{X}_{HL}}(i,j)+\hat{\mathbf{X}_{HH}}(i,j))/4, \\ 
&\hat{\mathbf{X}}(2i-1,2j)=(\hat{\mathbf{X}_{LL}}(i,j)-\hat{\mathbf{X}_{LH}}(i,j)\\
&\;\;\;\;\;\;\;\;\;\;\;\;\;\;\;\;\;\;\;\;\;\;\;\;\;\;\;\;\;+\hat{\mathbf{X}_{HL}}(i,j)-\hat{\mathbf{X}_{HH}}(i,j))/4, \\
&\hat{\mathbf{X}}(2i,2j-1)=(\hat{\mathbf{X}_{LL}}(i,j)+\hat{\mathbf{X}_{LH}}(i,j)\\
&\;\;\;\;\;\;\;\;\;\;\;\;\;\;\;\;\;\;\;\;\;\;\;\;\;\;\;\;\;-\hat{\mathbf{X}_{HL}}(i,j)-\hat{\mathbf{X}_{HH}}(i,j))/4, \\
&\hat{\mathbf{X}}(2i,2j)=(\hat{\mathbf{X}_{LL}}(i,j)+\hat{\mathbf{X}_{LH}}(i,j)\\
&\;\;\;\;\;\;\;\;\;\;\;\;\;\;\;\;\;\;\;\;\;\;\;\;\;\;\;\;\;+\hat{\mathbf{X}_{HL}}(i,j)+\hat{\mathbf{X}_{HH}}(i,j))/4,
\end{aligned}\right.
\label{IDWT}
\end{equation}
where $\hat{\mathbf{X}}(i,j)$ is the $(i,j)$-th index of $\hat{\mathbf{X}}$. Feature maps at different stages $\hat{\mathbf{X}} = \{\hat{\mathbf{X}^{1}},\hat{\mathbf{X}^{2}},\dots,\hat{\mathbf{X}^{i}}\}$ are obtained by Eq.~\eqref{IDWT}.

Benefiting from the IDWT, the multi-frequency subbands can be completely integrated into features $\hat{\mathbf{X}^{i}}$ that have the same dimensions as the input features $\mathbf{X}^{i}$. Moreover, the \textbf{\emph{correspondences}} between the low-frequency subband and the high-frequency subbands are established, thereby disentangling valuable structural information from the high- and low-frequency correspondences to achieve better real-world image denoising performance.

\subsection{Reconstruction}
The reconstruction phase employs three stages of RBs and a convolutional layer for feature reconstruction. The convolutional layers are adopted in the reconstruction phase for two reasons. 1) Although Transformer can be considered a specific instantiation of a spatially varying convolution~\cite{revisiting,scaling}, convolutional layers with spatially invariant filters can enhance the translational equivalence of DnSwin. 2) The RBs and concatenation operator aggregate the features at different levels.

We use $2 \times 2$ transposed convolutions with strides of 2 for the upsampling layers. The upsampling layers double the sizes of the feature maps and reduce the number of channels by half. After completing the upsampling operation, we concatenate $\hat{\mathbf{X}^{i-1}}$ and $\hat{\mathbf{X}^{i}}$ together to reduce the impact of the spatial information caused by downsampling. The denoised image $I_{Denoised}$ can be reconstructed by the following equation:
\begin{equation}
\left\{\begin{aligned}
    &\hat{\mathbf{X}^{i}}=H_{RB}(\hat{\mathbf{X}^{i}}), \\
    &\hat{\mathbf{X}^{i-1}}=H_{RB}(\operatorname{Concat}(\hat{\mathbf{X}^{i}} \uparrow,\hat{\mathbf{X}^{i-1}})), \\
    &\dots,\\
    &\hat{\mathbf{X}^{1}}=H_{RB}(\operatorname{Concat}(\hat{\mathbf{X}^{2}}\uparrow,\hat{\mathbf{X}^{1}})), \\
    &I_{Denoised}  = \operatorname{Conv}(\hat{\mathbf{X}^{1}}).
\end{aligned}\right.
\label{equ:decoder}
\end{equation}
where $\uparrow$ is the upsampling operation.

\subsection{Loss Function}
Instead of using the $\mathcal{L}_{1}$ loss, our model is optimized with the robust Charbonnier loss~\cite{Char} to better handle outliers and achieve improved performance. The Charbonnier loss is defined as follows:
\begin{equation}
    \mathcal{L}_{char} = \sqrt{\|I_{Denoised}-I_{GT}\|^{2}+\varepsilon^{2}}
    \label{equ:loss}
\end{equation}
where $I_{GT}$ represents the ground-truth image, and $\varepsilon$ is an empirical value. Compared with the $\mathcal{L}_{1}$ loss, the $\mathcal{L}_{char}$ loss makes the model more robust. In Section~\ref{sec:abltation}, we show that using the $\mathcal{L}_{char}$ loss is superior to using the $\mathcal{L}_{1}$ loss.

\section{Experiments}
\label{section: Experiment}

\begin{figure}[ht]
    \centering
    \begin{subfigure}{\textwidth}
        \centering
        \includegraphics[width=\textwidth]{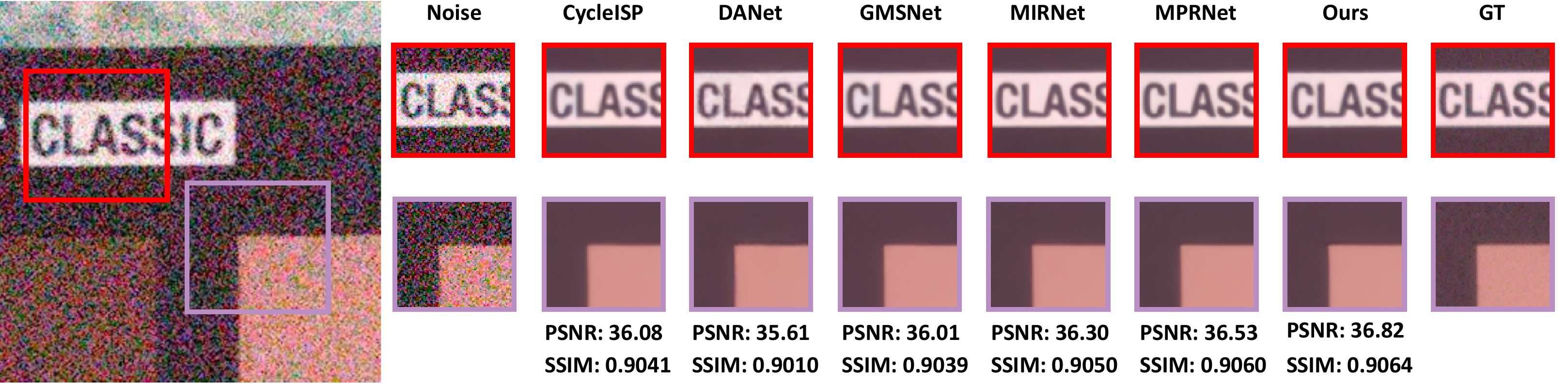}
    \end{subfigure}
    
    \vspace{0.5cm}
    \begin{subfigure}{\textwidth}
        \centering
        \includegraphics[width=\textwidth]{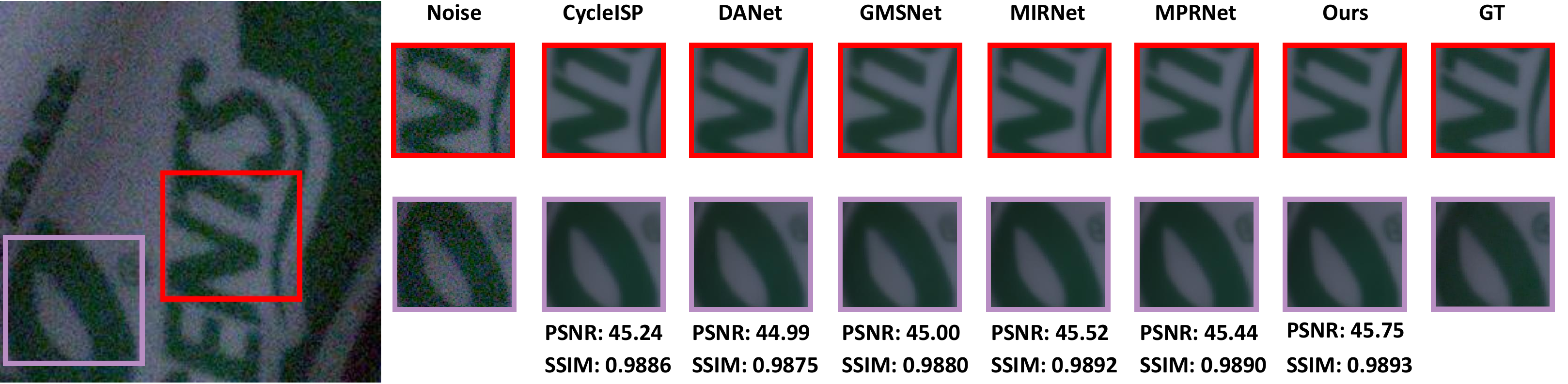}
    \end{subfigure}
    
    \vspace{0.5cm}
    \begin{subfigure}{\textwidth}
        \centering
        \includegraphics[width=\textwidth]{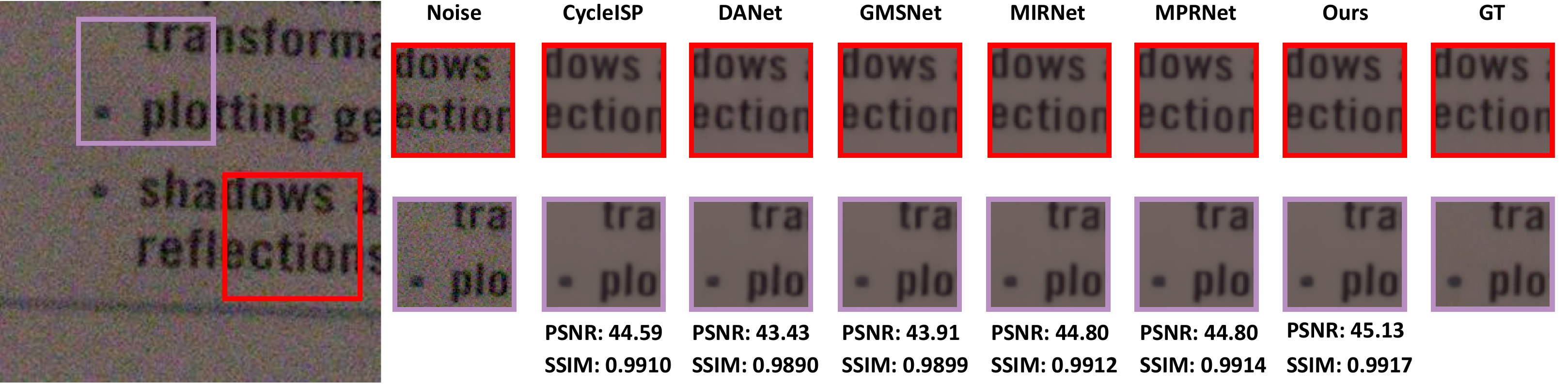}
    \end{subfigure}
    \caption{Comparison with state-of-the-art methods on real noisy images from the SIDD validation dataset~\cite{SIDD_2018_CVPR} (\textbf{Zoom in for the best view}).}
    \label{fig:sidd_val}
\end{figure}

\begin{figure}[ht]
    \centering
    \includegraphics[width=\textwidth]{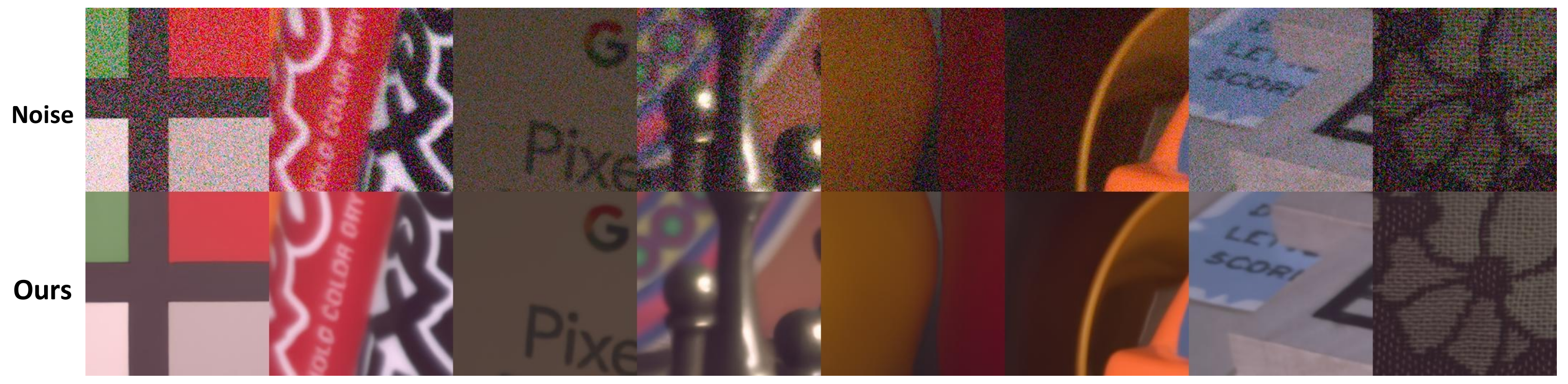}
    \caption{The visual results of DnSwin provided by SIDD benchmark online testing. Note that no ground-truth images or visual results of other state-of-the-art methods are included in the comparison (\textbf{Zoom in for the best view}).}
    \label{fig:sidd_benchmark}
\end{figure}

\begin{figure}[ht]
    \centering
    \begin{subfigure}{\textwidth}
        \centering
        \includegraphics[width=\textwidth]{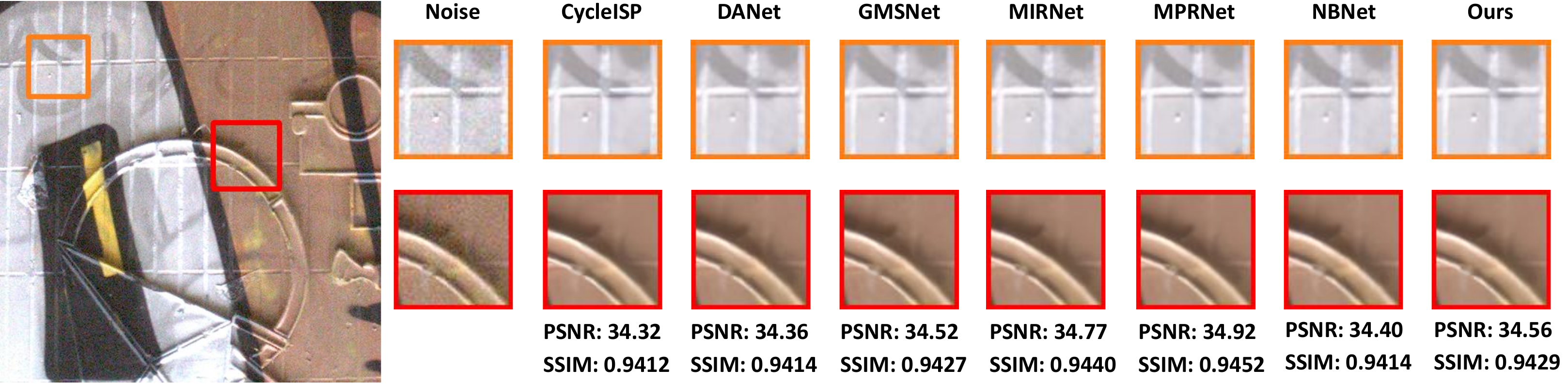}
    \end{subfigure}
    
    \vspace{0.5cm}
    \begin{subfigure}{\textwidth}
        \centering
        \includegraphics[width=\textwidth]{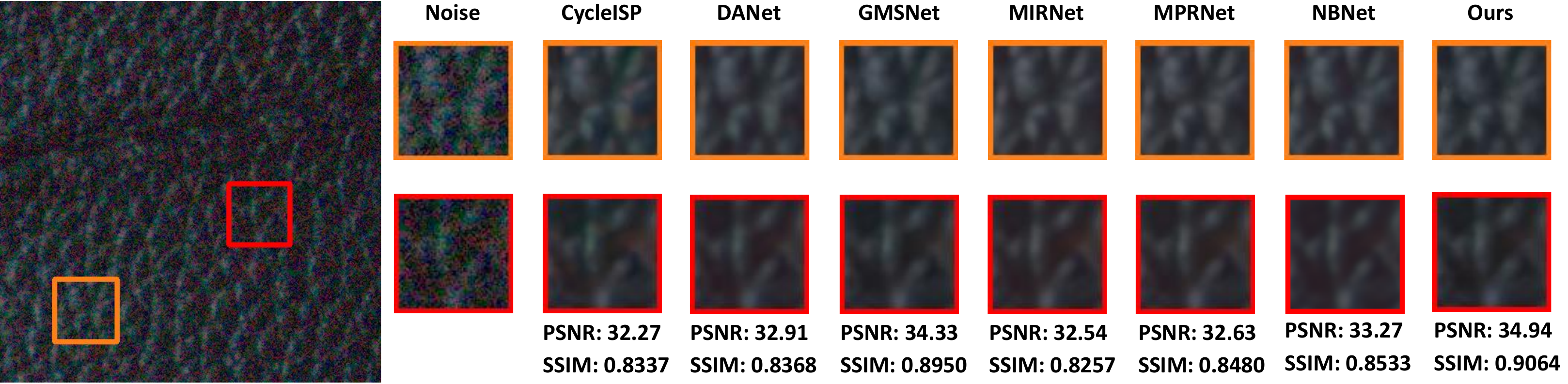}
    \end{subfigure}
    
    \vspace{0.5cm}
    \begin{subfigure}{\textwidth}
        \centering
        \includegraphics[width=\textwidth]{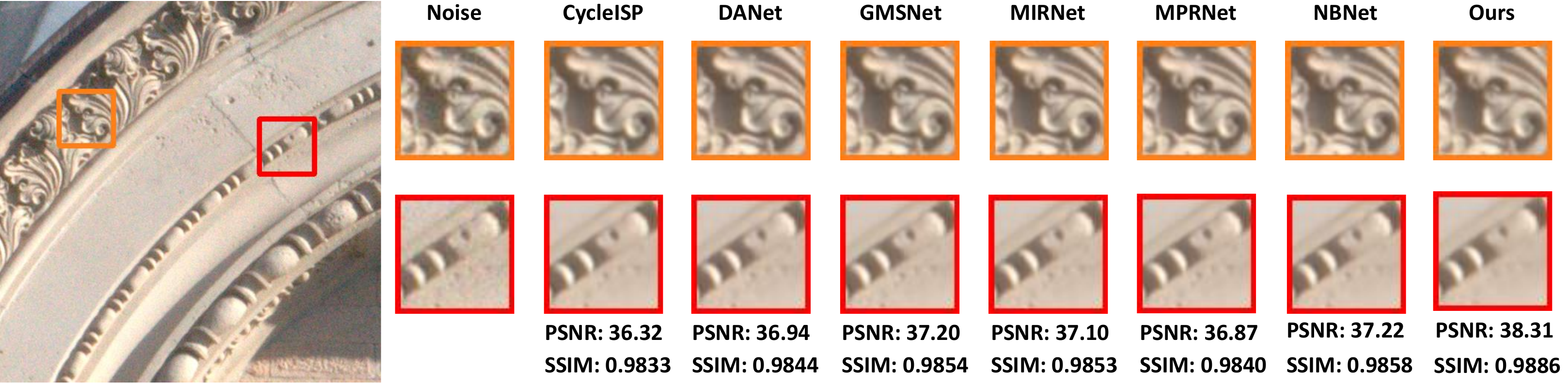}
    \end{subfigure}
    \caption{Comparison with state-of-the-art methods on real noisy images from the DND benchmark~\cite{dnd}. Notice that no ground-truth images are provided by the DND benchmark (\textbf{Zoom in for the best view}).}
    \label{fig:dnd}
\end{figure}

In this section, we first describe the experimental settings, including the training and testing settings and implementation details. Then, we compare our model with state-of-the-art methods on three real-world image denoising benchmarks (\textit{i.e.}, the SIDD validation dataset, SIDD benchmark and DND benchmark). Finally, we evaluate the contributions of different DnSwin components and parameters through an ablation study and analyze the efficiency of our model.
\subsection{Dataset and Evaluation Protocols}
We use the following two real-world noisy-clean datasets for conducting comprehensive comparisons to validate our model:

\begin{itemize}
  \item \emph{SIDD}~\cite{SIDD_2018_CVPR} consists of 30000 LR-HR image pairs, which were collected under different lighting conditions using five representative smartphone cameras, and their generated ground-truth images. For fast training and evaluation, we use the SIDD-Medium dataset (320 image pairs) for training and evaluate the approaches on the SIDD validation set and SIDD benchmark\footnotemark[1].
 
  \item \emph{DND}~\cite{dnd} consists of 50 pairs of real noisy images and their corresponding ground-truth images, which were captured with consumer-grade cameras possessing different sensor sizes. However, the DND\footnotemark[2] does not provide any additional training data for fine-tuning denoising networks and can only be evaluated online.
\end{itemize}

\footnotetext[1]{https://www.eecs.yorku.ca/\~kamel/sidd/benchmark.php}
\footnotetext[2]{https://noise.visinf.tu-darmstadt.de/benchmark/\#results\_srgb}

In the evaluation protocols, we adopt the PSNR and SSIM to verify our model because they are the most commonly used evaluation metrics in image restoration, as they focus on pixel fidelity.

\subsection{Implementation Details and Competing Methods}
Following the common training of GMSNet~\cite{gmsnet}, we use the Adam optimizer~\cite{adam} with $\beta_1 = 0.9$ and $\beta_2 = 0.999$ to train our framework. To improve the tradeoff between the size of the input patches and the available computing power, we set the batch size to 16 and the image patch size to $128 \times 128$. The initial learning rate is $2 \times 10^{-4}$. We employ the cosine decay strategy to decrease the learning rate to $1 \times 10^{-6}$. For all experiments, we use flipping and random rotation with angles of $90^{\circ}$, $180^{\circ}$ and $270^{\circ}$ for data augmentation. For the LF-WSA and HF-SWSA blocks, we set the window size $M=8$. All the experiments are carried out in a Linux environment with PyTorch (1.4.0) running on a server with 2 NVIDIA Tesla V100 GPUs. Nvidia CUDA 10.2 and cuDNN are utilized to accelerate the GPU computations. We utilize 6000 training epochs for DnSwin, which take approximately four days.

\begin{table}[ht]
\centering
\begin{tabular}{l|ccc}
\toprule
\multicolumn{1}{c|}{Methods}   & \multicolumn{1}{l}{PSNR$\uparrow$}   & \multicolumn{1}{l}{SSIM$\uparrow$}   & \multicolumn{1}{l}{Parameters} \\ \midrule \midrule
CBDNet~\cite{CBDNet} & 30.78 & 0.801 &4.31M\\
VDN~\cite{vdn} & 39.28 & 0.956 &7.81M \\
DANet~\cite{danet}        & 39.47 & 0.957 &63.01M \\
MIRNet~\cite{MIRNet} & 39.72   & 0.959  &31.18M\\
CycleISP~\cite{CycleISP}     & 39.52    & 0.957 &- \\
MPRNet~\cite{MPRNet}  & 39.71  & 0.958 &20.42M \\
NBNet~\cite{nbnet}        & 39.75 & \underline{0.959} &13.31M\\
GMSNet~\cite{gmsnet} & 39.63 & 0.956 &32.54M\\ 
Uformer~\cite{wang2021uformer} &\underline{39.77} & \underline{0.959} & 20.63M \\
SwinIR~\cite{liang2021swinir} &\underline{39.77} &0.958 &11.85M \\\midrule \midrule
Ours     & \textbf{39.80}  & \textbf{0.960}  &23.24M\\ \bottomrule
\end{tabular}
\caption{Quantitative results on the SIDD validation dataset~\cite{SIDD_2018_CVPR}.}
\label{tab:sidd-val}
\end{table}

\begin{table}[ht]
\centering
\begin{tabular}{l|ccc}
\toprule
\multicolumn{1}{c|}{Methods}   & \multicolumn{1}{l}{PSNR$\uparrow$}   & \multicolumn{1}{l}{SSIM$\uparrow$}   & \multicolumn{1}{l}{Parameters} \\ \midrule \midrule
CBDNet~\cite{CBDNet} & 33.28 & 0.868 &4.31M \\
VDN~\cite{vdn} & 39.26 & 0.955 &7.81M\\
DANet~\cite{danet}    & 39.43 & 0.956 &63.01M \\
GMSNet~\cite{gmsnet} & \underline{39.51} & \underline{0.958} &32.54M\\ \midrule \midrule
Ours     & \textbf{39.65}  & \textbf{0.959} &23.24M\\ \bottomrule
\end{tabular}
\caption{Quantitative results on the SIDD benchmark~\cite{SIDD_2018_CVPR}.}
\label{tab:sidd-benchmark}
\end{table}

To prove the efficiency and superiority of DnSwin, we compare it with state-of-the-art approaches on a real-world image denoising task. In particular, we evaluate its real-world image denoising performance on three test sets, the SIDD validation dataset~\cite{SIDD_2018_CVPR}, SIDD benchmark and DND benchmark.

\textbf{SIDD Validation Dataset and SIDD Benchmark.} For the SIDD validation dataset and SIDD benchmark, we follow the setting of DANet~\cite{danet}, which only uses the SIDD-Medium dataset for training. We compare our method with previously developed state-of-the-art methods, including CBDNet~\cite{CBDNet}, VDN~\cite{vdn}, DANet~\cite{danet}, MIRNet~\cite{MIRNet}, CycleISP~\cite{CycleISP}, MPNet~\cite{MPRNet}, NBNet~\cite{nbnet} and GMSNet~\cite{gmsnet}. It should be noted that no official test script is available for the corresponding test, so fewer baselines are used for comparison on the SIDD benchmark.

\textbf{DND.} For the DND benchmark, we train our model on the SIDD-Medium dataset and synthesize the noisy images provided by GMSNet~\cite{gmsnet} to conduct a fair comparison. The synthesized noisy images are generated by DIV2K~\cite{div2k}. The high-resolution images in DIV2K are cropped into nonoverlapping image patches, and then noise is added to these patches. We randomly sample 320 image patches from the SIDD-Medium dataset and 100 image patches from the synthetic noisy dataset for each training iteration. We use the same baselines as those implemented on the SIDD validation dataset for comparison purposes.

\begin{table}[ht]
\centering
\begin{tabular}{l|ccc}
\toprule
\multicolumn{1}{c|}{Methods}   & \multicolumn{1}{l}{PSNR$\uparrow$}   & \multicolumn{1}{l}{SSIM$\uparrow$}   & \multicolumn{1}{l}{Parameters} \\ \midrule \midrule
CBDNet~\cite{CBDNet} & 38.06 & 0.942 &4.31M \\
VDN~\cite{vdn} & 39.38 & 0.952 &7.81M \\
DANet~\cite{danet} & 39.58 & 0.955 &63.01M \\
MIRNet~\cite{MIRNet} & 39.88  & 0.956 &31.18M \\
CycleISP~\cite{CycleISP}     & 39.56    & 0.956   &- \\
MPRNet~\cite{MPRNet}  & 39.80  & 0.954 &20.42M\\
NBNet~\cite{nbnet}        & 39.89 & 0.955 &13.31M \\
GMSNet~\cite{gmsnet} & \underline{40.15} & \underline{0.961} &32.54M \\
Uformer~\cite{wang2021uformer} &39.96 & 0.956 & 20.63M  \\ 
SwinIR~\cite{liang2021swinir} &40.01 &0.958 &11.85M \\ \midrule \midrule
Ours     & \textbf{40.29}  & \textbf{0.962} &23.24M\\ \bottomrule
\end{tabular}
\caption{Quantitative results on the DND benchmark~\cite{dnd}.}
\label{tab:dnd}
\end{table}

\begin{table}[ht]
\centering{
\begin{tabular}{l|ccccc}
\toprule
\multicolumn{1}{c|}{Method}  & VDN & DANet & MIRNet & SwinIR & Ours         \\ \midrule \midrule
\multicolumn{1}{c|}{Time(frame/s)} & 0.1502 & \underline{0.0907} & 0.1524 &0.4726 & \textbf{0.0847} \\ \midrule
\multicolumn{1}{c|}{Parameters} & 7.81M & 63.01M & 31.78M &11.85M & 23.24M \\ \bottomrule
\end{tabular}}
\caption{Efficiency analysis for a 256$\times$256 image from the SIDD validation dataset~\cite{SIDD_2018_CVPR}.}
\label{tab:efficiency}
\end{table}

\subsection{Quantitative and Qualitative Comparisons}

\textbf{SIDD Validation Dataset.} As depicted in Table~\ref{tab:sidd-val}, the results indicate that our model achieves better performance than all the state-of-the-art methods. For instance, our method performs favorably against the recently developed state-of-the-art SwinIR method~\cite{liang2021swinir}, where the PSNR and SSIM of the proposed results are 0.03 dB and 0.001 higher than those of SwinIR, respectively, justifying the effectiveness of DnSwin. Compared with DANet~\cite{danet}, which uses a GAN~\cite{gan}, our model still achieves a 0.33 dB improvement in the PSNR index. Additionally, we present the visual comparison in \figurename~\ref{fig:sidd_val}. It can be observed that our model recovers sharper and clearer images than those of the other methods and is therefore more faithful to the ground truths, which verifies the effectiveness of the local dependencies captured by the CNN and the long-range pixel dependencies captured by the WSWT.

\textbf{SIDD Benchmark.} In Table~\ref{tab:sidd-benchmark}, we present the quantitative comparison results obtained on the SIDD benchmark. This shows that our model still achieves first place for all indices. Our model surpasses the prior state-of-the-art GMSNet method~\cite{gmsnet} by 0.14 dB and 0.001 in terms of the PSNR and SSIM, respectively. Some visual results obtained by DnSwin during online server testing are shown in \figurename~\ref{fig:sidd_benchmark}. The visual results show that DnSwin can not only successfully remove the noise but also preserve the texture details of images. It should be noted that the online server does not provide the ground truths and visual results of other state-of-the-art methods for comparison.

\textbf{DND Benchmark.} As depicted in Table~\ref{tab:dnd}, our model achieves a significant improvement over other state-of-the-art methods. Compared with MPRNet~\cite{MPRNet}, our model demonstrates performance gains of 0.49 dB and 0.008 in terms of the PSNR and SSIM, respectively, justifying the notion that building frequency correspondences via a sliding window makes our model more robust when handling information with different frequencies. Although DANet utilizes many more parameters, our model achieves a 0.71 dB gain in terms of the PSNR index. Compared with the ViT-based methods, Uformer~\cite{wang2021uformer} and SwinIR~\cite{liang2021swinir}, our model outperforms them by PSNRs margin of up to 0.30 dB. The qualitative results depicted in \figurename~\ref{fig:dnd} verify that our model achieves a significant visual quality improvement over CycleISP with a realistic texture and clear structure.

\textbf{Efficiency Analysis.} Despite achieving competitive results in quantitative comparisons, our model still exhibits outstanding running efficiency. The comparison in Table~\ref{tab:efficiency} shows that our model achieves a competitive efficiency level. Although VDN~\cite{vdn} has the fewest parameters, the noisy test image must be fed into two networks separately to obtain the final denoised image, which seriously reduces the inference speed of this method. Compared with MIRNet~\cite{MIRNet}, our model has nearly 2$\times$ the running efficiency and a significant restoration quality improvement. Compared with DANet~\cite{danet}, our model achieves an obvious quantitative improvement with a similar time consumption. Although SwinIR~\cite{liang2021swinir} has fewer parameters, it has the worst running efficiency due to the computational complexity of the self-attention mechanism.

\subsection{Ablation Study}
\label{sec:abltation}
In this section, we first conduct an ablation study on the `sliding mechanism', `Convolution $+$ Transformer' and `Wavelet $+$ Transformer' to justify the effectiveness of our proposed components. Moreover, we conduct additional ablation studies on the number of filters, window size, and loss function using the auxiliary training dataset. All the ablation experiments are evaluated on the DND benchmark.

\begin{table}[ht]
\centering
\begin{tabular}{c|ccc}
\toprule
\multicolumn{1}{c|}{Test Set} & \multicolumn{3}{c}{DND~\cite{dnd}}                                        \\ \midrule
\multicolumn{1}{c|}{Metric}   & \multicolumn{1}{l}{PSNR$\uparrow$}   & \multicolumn{1}{l}{SSIM$\uparrow$}   & \multicolumn{1}{l}{Parameters} \\ \midrule \midrule
Ours w/o LF-WSA   &40.08 &0.960 &22.8M \\
Ours w/o HF-SWSA    &40.03 &0.958 &22.7M \\ \midrule
Ours Full & \textbf{40.29} & \textbf{0.962} & 23.2M \\ \bottomrule
\end{tabular}
\caption{Ablation study on the effectiveness of the sliding mechanism.}
\label{Tab:effectiveness of sliding mechanism}
\end{table}

\begin{table}[ht]
\centering
\begin{tabular}{c|ccc}
\toprule
\multicolumn{1}{c|}{Test Set} & \multicolumn{3}{c}{DND~\cite{dnd}}                                        \\ \midrule
\multicolumn{1}{c|}{Metric}   & \multicolumn{1}{l}{PSNR$\uparrow$}   & \multicolumn{1}{l}{SSIM$\uparrow$}   & \multicolumn{1}{l}{Parameters} \\ \midrule \midrule
Standard UNet ($C=32$)  & 39.48  & 0.956 & 9.5M \\ \midrule
Ours ($C=16$) & 40.15 &0.957 &10.62M\\
Ours ($C=32$) & \textbf{40.29} & \textbf{0.962} & 23.2M \\ \bottomrule
\end{tabular}
\caption{Ablation study on the effectiveness of Convolution $+$ Transformer.}
\label{Tab:effectiveness of Convolution Transformer}
\end{table}

\begin{table}[ht]
\centering
\begin{tabular}{c|ccc}
\toprule
\multicolumn{1}{c|}{Test Set} & \multicolumn{3}{c}{DND~\cite{dnd}}                                        \\ \midrule
\multicolumn{1}{c|}{Metric}   & \multicolumn{1}{l}{PSNR$\uparrow$}   & \multicolumn{1}{l}{SSIM$\uparrow$}   & \multicolumn{1}{l}{Parameters} \\ \midrule \midrule
Standard UNet  & 39.48  & 0.956 & 9.5M \\
Ours w/o LF-WSA and HF-SWSA     & 39.97 &0.958 & 22.3M \\ 
Ours w/o DWT and IDWT    & 40.15 & 0.960 & 19.5M \\ \midrule 
Ours Full & \textbf{40.29} & \textbf{0.962} & 23.2M \\ \bottomrule
\end{tabular}
\caption{Ablation study on the effectiveness of Wavelet $+$ Transformer.}
\label{Tab:effectiveness of Wavelet $+$ Transformer.}
\end{table}

\begin{table}[ht]
\centering
\begin{tabular}{c|ccc}
\toprule
\multicolumn{1}{c|}{Test Set} & \multicolumn{3}{c}{DND~\cite{dnd}}                                        \\ \midrule
\multicolumn{1}{c|}{Metric}   & \multicolumn{1}{l}{PSNR$\uparrow$}   & \multicolumn{1}{l}{SSIM$\uparrow$}  & \multicolumn{1}{l}{Parameters} \\ \midrule \midrule
$C=16$  & 40.15  & 0.957 & 10.62M \\
$C=32$   & \textbf{40.29} & \textbf{0.962} & 23.24M \\ 
$C=44$   & 40.28  & 0.962 & 30.56M\\ 
$C=64$  &40.18 &0.960 &48.52M \\\bottomrule
\end{tabular}
\caption{Ablation study on the number of filters.}
\label{Tab:ablation_C}
\end{table}

\begin{table}[ht]
\centering{
\begin{tabular}{l|cc}
\toprule
\multicolumn{1}{c|}{Test Set} & \multicolumn{2}{c}{DND~\cite{dnd}}                                        \\ \midrule
\multicolumn{1}{c|}{Metric}   & \multicolumn{1}{l}{PSNR$\uparrow$}   & \multicolumn{1}{l}{Parameters}   \\ \midrule \midrule
$M=2$  & 39.52  & 8.81M \\
$M=4$   & 39.96 & 14.52M \\
$M=8$   & 40.29 & 23.24M \\
$M=16$   & \textbf{40.31} & 40.46M \\
\bottomrule
\end{tabular}}
\caption{Ablation study on the window size $M$.}
\label{Tab:ablation_windowsize}
\end{table}

\begin{table}[ht]
\centering
\begin{tabular}{l|cc}
\toprule
\multicolumn{1}{c|}{Test Set} & \multicolumn{2}{c}{DND~\cite{dnd}}                                        \\ \midrule
\multicolumn{1}{c|}{Metric}   & \multicolumn{1}{l}{PSNR$\uparrow$}   & \multicolumn{1}{l}{SSIM$\uparrow$}   \\ \midrule \midrule
$\mathcal{L}=\mathcal{L}_{1}$  & 40.06  & 0.958  \\
$\mathcal{L}=\mathcal{L}_{char}$   & \textbf{40.29} & \textbf{0.962} \\ \bottomrule
\end{tabular}
\caption{Ablation study on the loss function.}
\label{Tab:ablation_loss}
\end{table}

\begin{table}[t]
\centering
\begin{tabular}{c|cc}
\toprule
\multicolumn{1}{c|}{Test Set} & \multicolumn{2}{c}{DND~\cite{dnd}}                                        \\ \midrule
\multicolumn{1}{c|}{Metric}   & \multicolumn{1}{l}{PSNR$\uparrow$}   & \multicolumn{1}{l}{SSIM$\uparrow$}   \\ \midrule \midrule
$\mathcal{D}_{1}$: denoising dataset synthesized by DIV2K &39.42 &0.951 \\
$\mathcal{D}_{2}$: real denoising dataset SIDD-Medium  & 39.54  & 0.952  \\
$\mathcal{D}_{1}+\mathcal{D}_{2}$  & \textbf{40.29} & \textbf{0.962} \\ \bottomrule
\end{tabular}
\caption{Ablation study on the auxiliary training dataset.}
\label{Tab:ablation_data}
\end{table}

\textbf{Effects of the Sliding Mechanism.} We conduct an ablation study on the sliding mechanism and present the experimental results. As depicted in Table~\ref{Tab:effectiveness of sliding mechanism}, `Ours w/o LF-WSA' achieves a 0.05 dB improvement, a 0.002 SSIM gain and a similar number of parameters in a comparison with `Ours w/o HF-SWSA', which justifies the effectiveness of the sliding mechanism in real-world denoising scenarios. Moreover, compared with `Ours w/o HF-SWSA', `Ours Full' demonstrates better performance with 0.26 dB and 0.004 SSIM improvements, further justifying the effectiveness of the sliding mechanism and the effectiveness of the correspondences between the multi-frequency features obtained by incorporating the LF-WSA and HF-SWSA.

\textbf{Convolution $+$ Transformer.} We conduct an analysis to verify the effectiveness of the convolutional operator and Transformer. Specifically, we replace all the LF-WSA and HF-SWSA blocks with ResBlocks~\cite{he2016deep} and remove the DWT and IDWT operations, resulting in `Standard UNet ($C=32$), as shown in Table~\ref{Tab:effectiveness of Convolution Transformer}. We observe that `Ours ($C=16$)' achieves a 40.15 dB PSNR and outperforms `Standard UNet ($C=32$)' by 0.67 dB and 0.001 in terms of the PSNR and SSIM, respectively, with a closed number of parameters, which verifies that the architecture with `Convolution $+$ Transformer' possesses a better local and global context modeling ability than the pure CNN-based model. In addition, `Ours ($C=32$)' surpasses `Standard UNet ($C=32$)' by 0.81 dB and 0.006 in terms of the PSNR and SSIM, respectively, with more parameters, which indicates the effectiveness of `Convolution $+$ Transformer' in real-world denoising. We define $C$ as the number of filters in the convolution layer in each model.

\textbf{Effects of Wavelet $+$ Transformer.} To analyze the effectiveness of Wavelet $+$ Transformer, we conduct an ablation study on different network architectures and observe their results obtained on the DND benchmark~\cite{dnd}. As shown in Table~\ref{Tab:effectiveness of Wavelet $+$ Transformer.}, compared with `Standard UNet', `Ours w/o LF-WSA and HF-SWSA' achieves limited a performance improvement with nearly $2.5 \times$ the parameters, which certainly justifies the notion that simply decomposing features using the DWT and integrating them using the IDWT is useless for real-world image denoising yet brings a significant parameter overhead. Additionally, `Ours Full' obviously surpasses `Ours w/o DWT and IDWT' by 0.24 dB and 0.002 in terms of the SSIM, which justifies the effectiveness of Wavelet $+$ Transformer. Moreover, `Ours Full' exhibits the best PSNR and SSIM results, benefitting from the combination of multi-frequency feature decomposition, integration and between-feature correspondence building. Therefore, we apply the Wavelet $+$ Transformer architecture in the proposed DnSwin.

\textbf{Number of Filters.} We conduct an experiment on the number of filters $C$. Table~\ref{Tab:ablation_C} shows that `$C=16$' achieves a slightly worse performance than that obtained with `$C=32$' but at a lower computational cost. In addition, `$C=44$' yields a similar performance to that of `$C=32$' with an obvious parameter increment. However, when `$C=64$', the performance becomes significantly worse because the large amount of redundant information introduced by the unnecessary filters may hamper the network performance. Although the performance can be improved with a larger number of filters $C$, the computational complexity also increases, so we choose a compromise number of filters $C=32$ to keep the computational complexity suitable while maintaining better performance.

\textbf{Window Size.} We perform an ablation study on the window size $M$. Due to patch boundary artifacts, the patch size (128) should be an integer multiple of the window size. As shown in Table~\ref{Tab:ablation_windowsize}, we choose $M=\{2,4,8,16\}$ to verify the effects of different window size settings. When `$M=2$' or `$M=4$', the window is too small to capture the long-range dependencies of pixels, which leads to a significant drop in the model performance. Compared with `$M=8$', `$M=16$' only achieves only a marginal performance improvement while significantly increasing the number of model parameters. Based on the above analysis, we choose $M=8$ as the default window size.

\textbf{Loss Function.} We compare different loss settings on the DND benchmark, as shown in Table~\ref{Tab:ablation_loss}. `$\mathcal{L}=\mathcal{L}_{char}$' achieves 40.29 dB, and `$\mathcal{L}=\mathcal{L}_{1}$' obtains 40.06 dB, justifying the effectiveness of the $L_{char}$ loss in real-world image denoising. Moreover, $L_{char}$ also yields a 0.004 SSIM gain over that of $L_{1}$. To this end, we replace $L_{1}$ with $L_{char}$ as the default loss function in the model optimization process.

\textbf{Auxiliary Training Dataset.} We also explore the impact of the auxiliary training dataset on the model performance. As depicted in Table~\ref{Tab:ablation_data}, the results show that using the synthesized dataset (i.e., DIV2K $\mathcal{D}_{1}$) or the real-world denoising dataset (SIDD-Medium $\mathcal{D}_{2}$) leads to lower model performance. Synthetic denoising datasets are more important than real denoising datasets. Abdelhamed et al.~\cite{SIDD_2018_CVPR} only collected image pairs in a few indoor scenarios, which makes the trained model overfit these scenarios. However, our model achieves the best score when it is trained on $\mathcal{D}_{1}+\mathcal{D}_{2}$, which verifies that auxiliary training datasets can effectively improve the real-world image denoising performance of the model.

\section{Conclusion and Future Work}
\label{section: Conclusion}
In this paper, we propose an effective real-world denoising ViT called DnSwin by incorporating a WSWT. We integrate the advantages of both a CNN and the ViT, not only taking advantage of the CNN to process images with large sizes due to its local receptive field but also utilizing a self-attention mechanism to capture long-range dependencies. Moreover, with wavelet feature decomposition, by building frequency correspondences between multi-frequency subbands and multi-frequency integration, our model can handle features with different frequencies, especially high-frequency noise features, and produces promising real-world image denoising results. Experimental results on real-world denoising benchmarks demonstrate that our model greatly surpasses the previous state-of-the-art methods with a suitable number of parameters and faster inference times on Euclidean-based evaluation protocols.

We aim to extend our work in the following directions. First, since real-world images have more unknown degradation, we would like to change our model to a GAN~\cite{gan} model so that we can obtain results with better perceptual quality. Second, we are considering extending DnSwin to handle more general restoration tasks, such as image deblurring, deraining, and super-resolution.

\section*{Acknowledgments}
This work is supported in part by the National Natural Science Foundation of China (no. 62002069), Science and Technology Project of Guangdong Province (no. 2021A1515011341), Guangzhou Science and Technology Plan Project (no. 202002030386), and Guangdong Provincial Key Laboratory of Human Digital Twin (2022B1212010004).

 \bibliographystyle{elsarticle-num} 
 \bibliography{cas-refs}

\end{document}